\documentclass{article}
\PassOptionsToPackage{prologue,dvipsnames}{xcolor}
\pdfoutput=1
\usepackage{tikz}
\usepackage{subcaption}
\usepackage{soul}
\usepackage[dvipsnames]{xcolor}
\usepackage[square,numbers]{natbib}
\usepackage[preprint]{2024}
\usepackage[utf8]{inputenc} 
\usepackage[T1]{fontenc}    
\usepackage[pagebackref=true,breaklinks=true,colorlinks,bookmarks=false]{hyperref}
\usepackage{url}            
\usepackage{booktabs}       
\usepackage{amsfonts}       
\usepackage{nicefrac}       
\usepackage{microtype}      
\usepackage{graphicx, caption, wrapfig} 
\usepackage{amsmath}
\usepackage{physics}
\usepackage{pifont}
\newcommand{\cmark}{\ding{51}}
\newcommand{\xmark}{\ding{55}}
\newcommand{\yes}{\textcolor{Green}{\cmark}}
\newcommand{\no}{\textcolor{red}{\xmark}}

\title{TexVerse: A Universe of 3D Objects with High-Resolution Textures}

\author{%
  Yibo Zhang$^{1,2}$ \qquad
  Li Zhang$^{1, 3}$ \qquad
  Rui Ma$^{2}$\thanks{Corresponding author.} \qquad 
  Nan Cao$^{1, 4}$ \qquad
  \\
  \vspace{-0.2cm}
  \\ 
  $^1$Shanghai Innovation Institute \quad
  $^2$Jilin University \quad
  $^3$Fudan University \quad
  $^4$Tongji University \quad 
  \vspace{.5em} 
  \\ 
  \url{https://github.com/yiboz2001/TexVerse}
}

\usepackage{hyperref}
\usepackage{cleveref}
\crefname{figure}{fig.}{figures}
\Crefname{figure}{Fig.}{Figures}
\crefname{equation}{Eq.}{equations}
\crefname{section}{Sec.}{sections}
\Crefname{Section}{Sec.}{Sections}

\crefname{appendix}{appendix}{appendixs}
\Crefname{Appendix}{Appendix}{Appendixs}

\usepackage{soul}
\usepackage{float}
\usepackage{enumitem}

\usepackage{colortbl} 
\usepackage{multirow}
\usepackage{svg}
\usepackage{calc}
\usepackage[ruled]{algorithm2e} 

\SetAlFnt{\small}
\SetAlCapFnt{\small}
\SetAlCapNameFnt{\small}
\SetAlCapHSkip{0pt}

\begin{document}
\vspace*{-1cm}
\maketitle
\begin{figure*}[h]
    \centering
    \vspace{-23pt}
    \includegraphics[width=1\linewidth]{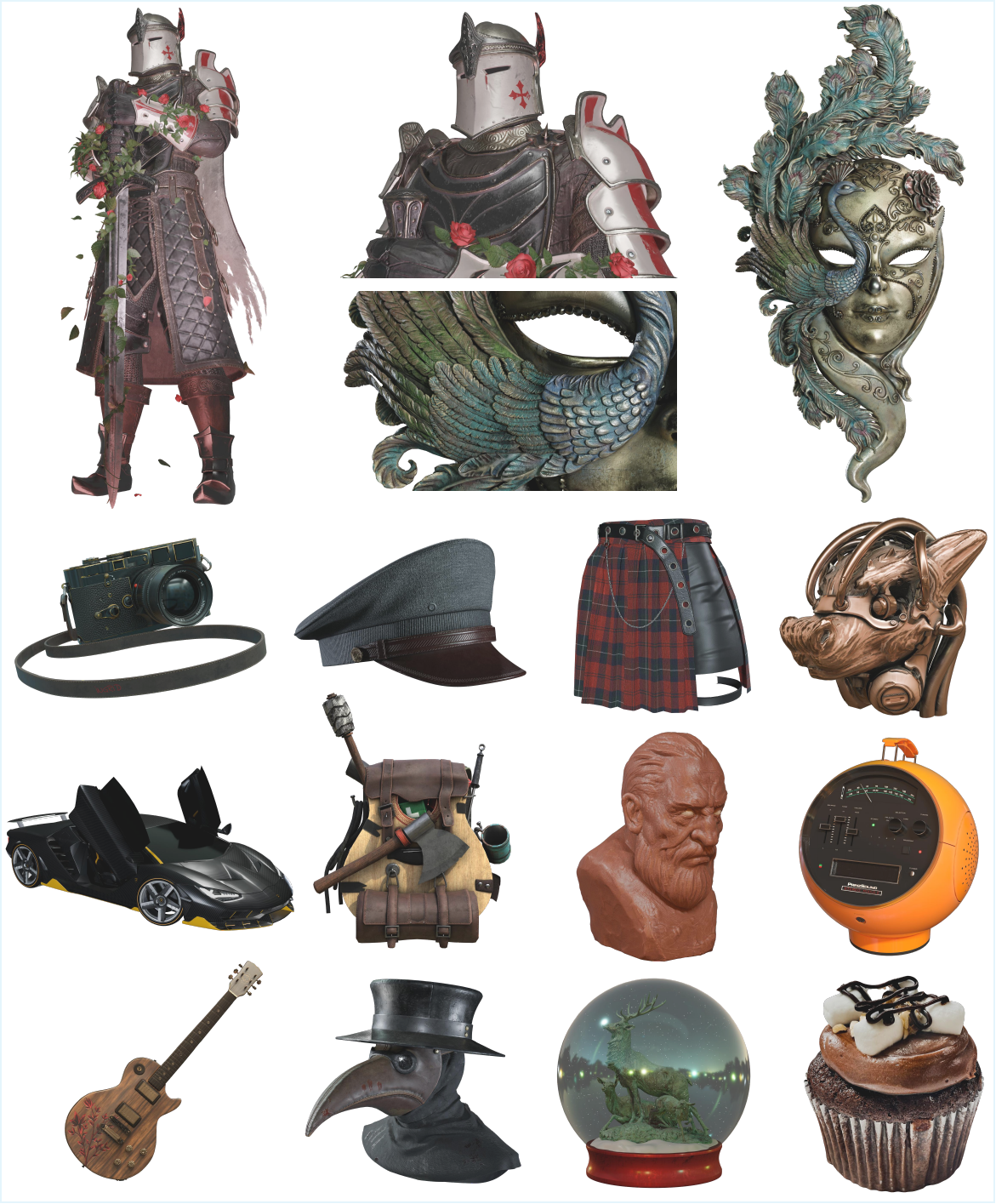}
    \caption{We introduce TexVerse, a large-scale 3D asset dataset featuring high-resolution textures. 
    }
    \label{fig:teaser}
\end{figure*}
\clearpage


\begin{abstract}
We introduce TexVerse, a large-scale 3D dataset featuring high-resolution textures. While recent advances in large-scale 3D datasets have enhanced high-resolution geometry generation, creating high-resolution textures end-to-end remains underexplored due to the lack of suitable datasets. TexVerse fills this gap with a curated collection of over 858K unique high-resolution 3D models sourced from Sketchfab, including more than 158K models with physically based rendering (PBR) materials. Each model encompasses all of its high-resolution variants, bringing the total to 1.6M 3D instances. TexVerse also includes specialized subsets: TexVerse-Skeleton, with 69K rigged models, and TexVerse-Animation, with 54K animated models, both preserving original skeleton and animation data uploaded by the user. We also provide detailed model annotations describing overall characteristics, structural components, and intricate features. TexVerse offers a high-quality data resource with wide-ranging potential applications in texture synthesis, PBR material development, animation, and various 3D vision and graphics tasks.
\end{abstract}
\section{Introduction}
Three-dimensional (3D) digital assets have become deeply integrated into modern life and industry, permeating fields from gaming and film to embodied artificial intelligence. However, creating high-quality 3D assets is often complex, time-consuming, and costly, requiring precise mesh structures, high-resolution textures, and reliance on specialized skills and intricate toolchains. As the demand for 3D digital experiences continues to grow, the automated generation of high-quality 3D assets has emerged as a key focus for both industry and academia in recent years.

Benefiting from the release of large-scale 3D datasets \cite{Objaverse-XL, Objaverse}, 3D generation techniques have advanced rapidly in recent years. 
Current state-of-the-art methods \cite{ultra3d, SparseFlex, Hunyuan3D_2.5, Sparc3D, Direct3D-S2} have achieved remarkable breakthroughs in the domain of high-resolution geometry generation.
However, in the field of texture and physically based rendering (PBR) material generation, existing approaches \cite{MVPaint, Unique3D, CLAY, Hunyuan3D_2.0} still rely on generating low-resolution results followed by post-processing techniques, such as super-resolution, to enhance resolution and quality. The capability for end-to-end generation of high-resolution textures and PBR materials remains significantly underexplored. 

This observation has sparked our interest. We recognize a clear gap in the community regarding the availability of accessible, large-scale, high-resolution texture datasets for 3D objects.
Currently, the only open-source datasets with sample sizes exceeding 100K are Objaverse (818K) and Objaverse-XL (10.2M). 
Taking Objaverse as an example, its models are sourced from artist creations on Sketchfab \cite{Sketchfab}, yet roughly half of the objects have textures with a maximum resolution below 1024 pixels or lack textures entirely. 
\textbf{Even for objects labeled in the original metadata as having higher-resolution textures (e.g., 4096), Objaverse only provides versions limited to 1024 resolution}, presenting a significant constraint. 
While Objaverse-XL is vastly larger in scale, its data, primarily sourced from the web (e.g., GitHub), exhibit considerable heterogeneity in quality and pose substantial challenges for data cleaning. 
The recently introduced Digital Twin Catalog (DTC) dataset \cite{DTC}, which includes approximately 2,000 high-precision scanned 3D objects with 4K PBR materials, offers superior quality but lacks the scale required to support the data-driven generation demand.

To fill this gap, we introduce \textbf{TexVerse}, a large-scale 3D dataset featuring high-resolution textures. 
Sketchfab hosts approximately 1.6 million freely downloadable 3D models. 
We curate TexVerse by filtering models with texture resolutions of at least 1024 pixels, excluding those tagged or described with terms related to ``NoAI'' and retaining only models licensed under distributable Creative Commons licenses. 
TexVerse encompasses \textbf{858,669} distinct high-resolution 3D models, of which \textbf{158,518} incorporate PBR materials, all standardized in the \texttt{.glb} format. 
Additionally, for each model, we also collect all of its high-resolution variants (e.g., the 4096 and 1024 versions of a model with a maximum resolution of 8192), yielding a total of \textbf{1,661,101} 3D instances. 
For the rigged and animated categories of models, we further obtain the original user-uploaded file format to prevent the loss of skeletons and animations during the format conversion of Sketchfab. 
These are named as the \textbf{TexVerse-Skeleton} and \textbf{TexVerse-Animation} datasets, comprising 69,138 rigged models and 54,430 animated models, respectively, all with high-resolution textures.
Furthermore, we provide \textbf{856,312} detailed annotations to the model generated by GPT-5 \cite{GPT-5}, which encompass the general description, structural composition and detailed characteristics.
We believe TexVerse will drive the community forward in areas such as high-resolution texture generation, PBR material synthesis, animation, and a wide range of 3D vision and graphics tasks.
\section{Related work}
\begin{table*}[t]
\centering
\newcommand{\synthetic}{\textcolor{red}{synthetic}}
\newcommand{\studio}{\textcolor{red}{studio}}
\newcommand{\wild}{\textcolor{Green}{in-the-wild}}
\caption{Comparison with commonly used 3D datasets.
Objaverse~\cite{Objaverse}, Objaverse-XL\cite{Objaverse-XL} and TexVerse consist of both synthetic objects and real scans, with only a subset containing PBR materials.
Within Objaverse and Objaverse-XL, only a subset provide high-resolution textures.
}

\tiny
\resizebox{\linewidth}{!}{
\begin{tabular}{lcccc}
\toprule


Dataset & \# Objects & Type & PBR Material & High-Resolution Texture \\
\midrule
ShapeNet~\cite{ShapeNet} & 51K & synthetic & \no & \no \\
3D-Future~\cite{3D-FUTURE} & 10K & synthetic & \no & \yes \\
ABO~\cite{ABO} & 8K & synthetic & \yes & \yes \\
OmniObject3D~\cite{OmniObject3D} & 6K & real & \no & \no \\
GSO~\cite{GSO} & 1K & real & \no & \yes \\
DTC~\cite{DTC} & 2K & real & \yes & \yes \\
Objaverse~\cite{Objaverse} & 818K & both & (\yes)* & (\yes)* \\
Objaverse-XL~\cite{Objaverse-XL} & 10M & both & (\yes)* & (\yes)* \\
TexVerse (Ours) & 858K & both & (\yes)* & \yes \\
\bottomrule
\end{tabular}

}
\label{tab:datasets}
\end{table*}
We provide a comparison of our TexVerse dataset to existing commonly used 3D datasets in \Cref{tab:datasets}.
Synthetic datasets constitute a significant portion of existing resources. 
ShapeNet \cite{ShapeNet} comprises approximately 51,000 mesh models with intricate geometric structures but limited texture resolution. Likewise, 3D-FUTURE \cite{3D-FUTURE} and ABO \cite{ABO} focus on furniture and consumer goods, though their scales remain relatively modest. Objaverse \cite{Objaverse} includes around 818,000 artist-created 3D models, primarily sourced from Sketchfab. However, nearly half of these models suffer from low-resolution textures, with some lacking textures entirely, and even those labeled as high-resolution are limited to 1024 resolution, restricting their use in tasks requiring high precision. 
Its expanded version, Objaverse-XL \cite{Objaverse-XL}, encompasses 10 million objects from web (e.g., GitHub), but its significant quality heterogeneity presents substantial data cleaning challenges.
In contrast, real-world datasets, limited by the difficulties of acquiring high-quality 3D data in the wild, are typically smaller in scale. For instance, GSO \cite{GSO}, OmniObject3D \cite{OmniObject3D}, and the recent DTC \cite{DTC} offer high-quality scanned models but are restricted to a few thousand objects, insufficient for large-scale data-driven applications. 
As presented in \Cref{tab:datasets}, our TexVerse dataset, comprising 858,669 objects, includes both synthetic objects and real scan. All objects feature high-resolution textures, with a subset incorporating PBR materials, providing strong support for advanced 3D research.
\section{TexVerse}
TexVerse is a large-scale 3D dataset featuring high-resolution textures. 
The objects are sourced from Sketchfab, an online 3D marketplace where users can upload and share models for both free and commercial use.
We conducted a comprehensive survey of Sketchfab, identifying approximately 1.6 million freely downloadable 3D models uploaded between 2012 and 2025. Using metadata provided by Sketchfab, we first filtered for high-resolution textured models with texture resolutions of at least 1024 pixels, \textbf{excluding models tagged or described with terms related to ``NoAI''}.  
Then, we obtained models which under the distributable Creative Commons license using Sketchfab’s API. 
The resulting dataset, formatted in the \texttt{.glb} format, comprises \textbf{858,669} unique high-resolution textured 3D objects across various resolution levels. 
For each model, we also collected all of its high-resolution variants (e.g., the 4096 and 1024 versions of a model with a maximum resolution of 8192), yielding a total of \textbf{1,661,101} 3D instances.

\subsection{Metadata}
\label{sec:dataset_metadata}

For each object in the TexVerse, we provide metadata extracted from the information supplied by its creator upon uploading to Sketchfab: \texttt{uid}, \texttt{name}, \texttt{description}, 
\texttt{user-name}, \texttt{tags}, \texttt{categories}, \texttt{thumbnail-url}, \texttt{vertex-count}, \texttt{face-count}, \texttt{max-texture}, \texttt{pbr-type}, \texttt{is-rigged}, \texttt{animation-count}, \texttt{license}.

\begin{figure*}[t]
    \centering
    \includegraphics[width=1\textwidth]{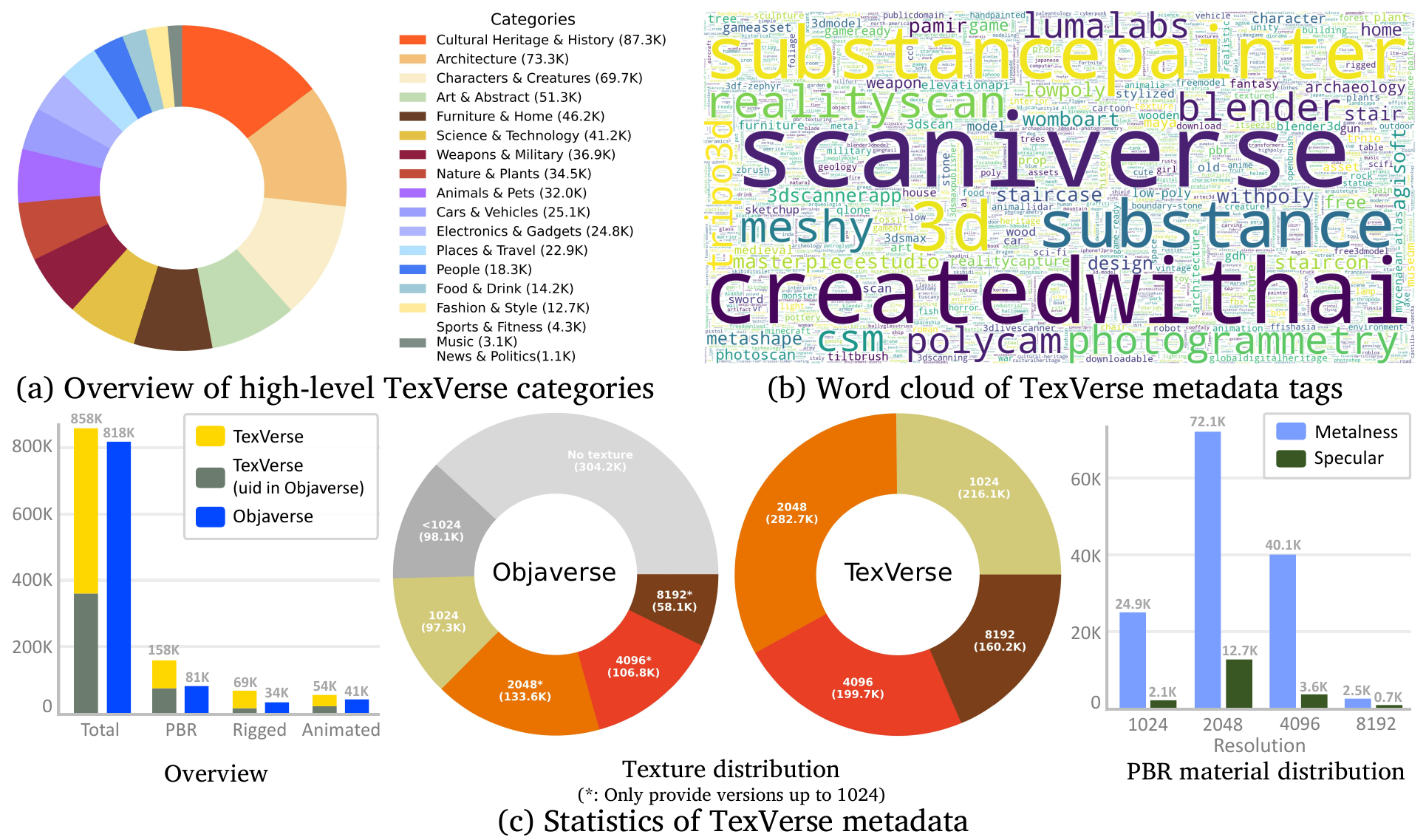}
    \caption{\textbf{TexVerse statistics.}
    (a) Distribution of high-level TexVerse categories; (b) Word cloud of metadata tags; (c) Metadata statistics, including metadata comparison between TexVerse and Objaverse, texture resolution distribution, and PBR material distribution.}
    \label{fig:statistics}
\end{figure*}
\subsection{Statistics}
\label{sec:dataset_statistics}

We conduct detailed analyses of TexVerse, as presented in \Cref{fig:statistics}.
\Cref{fig:statistics}(a) illustrates the category distribution, (b) displays the word cloud of tags, and (c) offers a static analysis of critical metadata alongside the comparison with Objaverse dataset.

\begin{figure*}[t]
    \centering
    \includegraphics[width=1\linewidth]{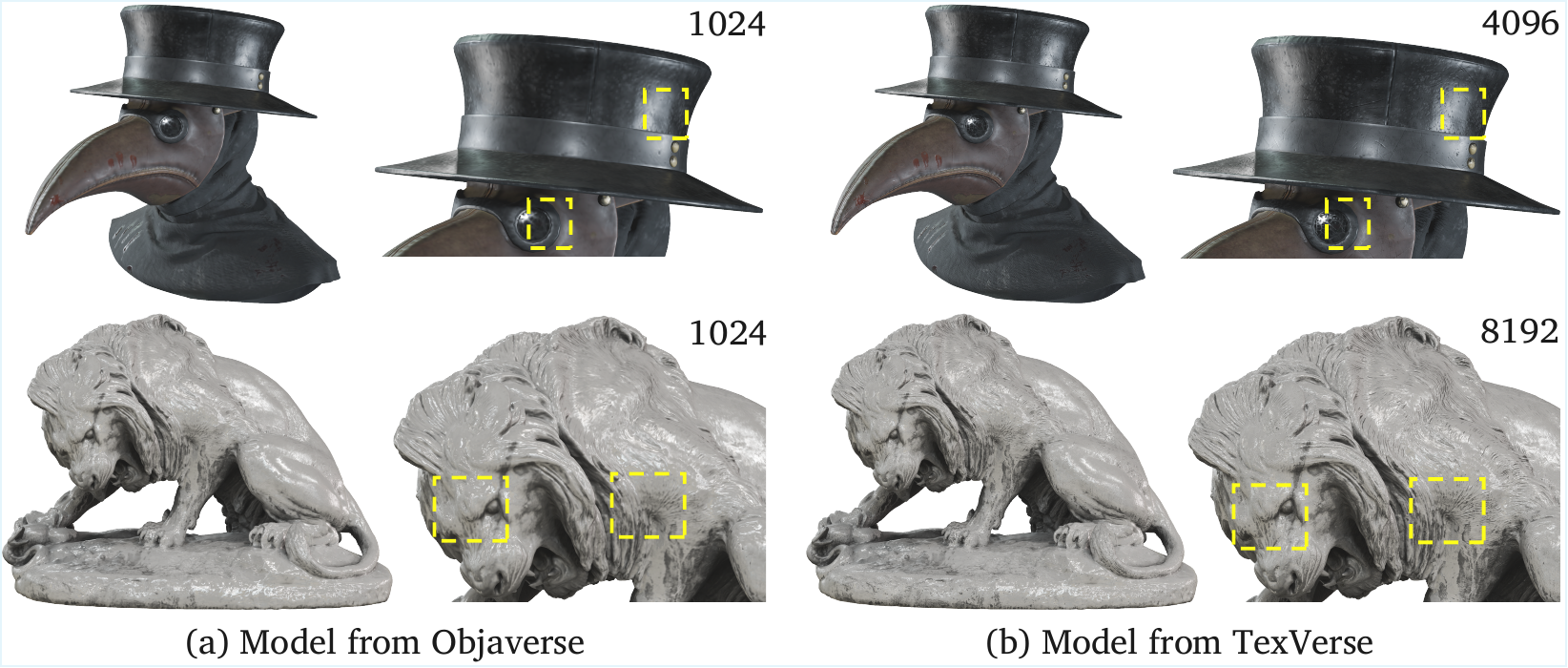}
    \caption{Objaverse only provides versions up to 1024 resolution for objects labeled as having higher-resolution textures in the metadata, whereas we provide genuine high-resolution versions.
    The UIDs are \texttt{d4d12479b5bb4bfaa72dbcf1955d5eb7} and \texttt{d5e6b6a11da646f68a5fcba661dcae99}.
    }
    \label{fig:compare}
\end{figure*}

\paragraph{Compared with Objaverse dataset}
The Objaverse dataset, similar to ours, consists of 818K Sketchfab models uploaded between 2012 and 2022.
As shown in \Cref{fig:statistics}(c, middle), TexVerse demonstrates clear superiority in terms of high-resolution textures.
Approximately half of the objects in Objaverse have textures with a maximum resolution below 1024 pixels, or lack textures entirely.
\textbf{Even for objects labeled in the original metadata as having higher-resolution textures (e.g., 4096), Objaverse only provides versions limited to 1024 resolution.
}
We provide some examples in \Cref{fig:compare}.
In contrast, TexVerse significantly outperforms Objaverse across all high-resolution texture levels.
In addition, \Cref{fig:statistics}(c, left) shows that TexVerse comprises over 858,000 objects, with nearly $60\%$ representing novel models not present in Objaverse.
It also surpasses Objaverse in the amount of critical metadata, underscoring its potential to advance a wide range of downstream tasks, such as geometry generation and skeleton generation tasks.

\paragraph{Rigged and animated models.}
TexVerse comprises 69,138 rigged models (i.e., models with a digital skeleton used to animate 3D models) and 54,430 animated models, maintaining high-resolution textures.
However, we observe that during Sketchfab’s conversion of user-uploaded raw format into the standardized \texttt{.glb} format, essential information such as skeletons or animations is often discarded.
To mitigate this issue, we also acquire the original file formats uploaded by users for rigged and animated models, which we designate as \textbf{TexVerse-Skeleton} and \textbf{TexVerse-Animation}.

\begin{figure*}[t]
    \centering
    \includegraphics[width=1\linewidth]{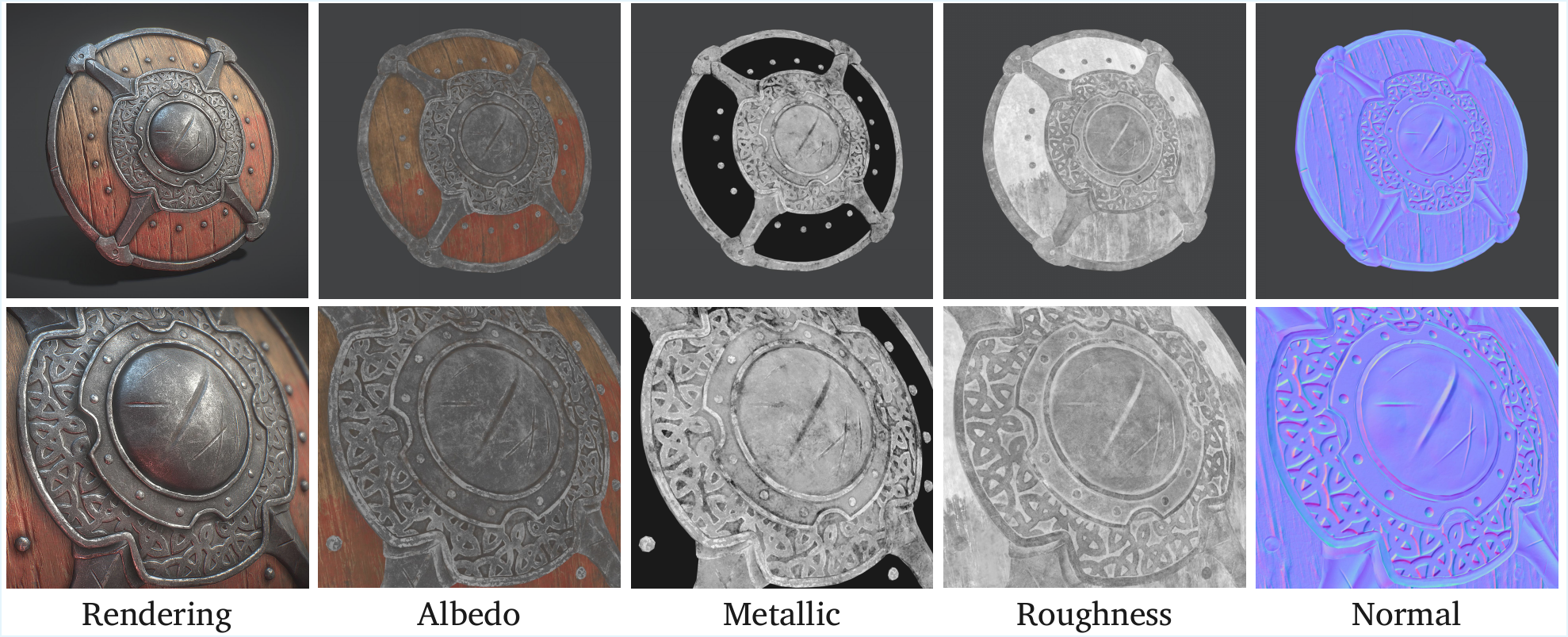}
    \caption{Data example with 4K resolution PBR materials. Please zoom in to check the detail.}
    \label{fig:pbr}
\end{figure*}

\paragraph{PBR material}
TexVerse comprises 158,518 models with high-resolution physically based rendering (PBR) materials, following two standard workflows: \texttt{Metalness} and \texttt{Specular}.
To qualify as PBR, each material must include a texture in the roughness or glossiness channel, as well as in either the metalness or specular channel, depending on the chosen workflow.
The distribution of PBR materials across different texture resolutions is shown in \Cref{fig:statistics}(c, right).
We provide an example featuring 4K resolution PBR materials, as shown in \Cref{fig:pbr}.

\paragraph{License}
All models in our dataset are released under distributable Creative Commons licenses. Over $80\%$ of models are under \texttt{CC BY} or \texttt{CC0} licenses, enabling flexible use in both academic and commercial settings.


\subsection{Model annotation}
\begin{figure*}[t]
    \centering
    \includegraphics[width=1\linewidth]{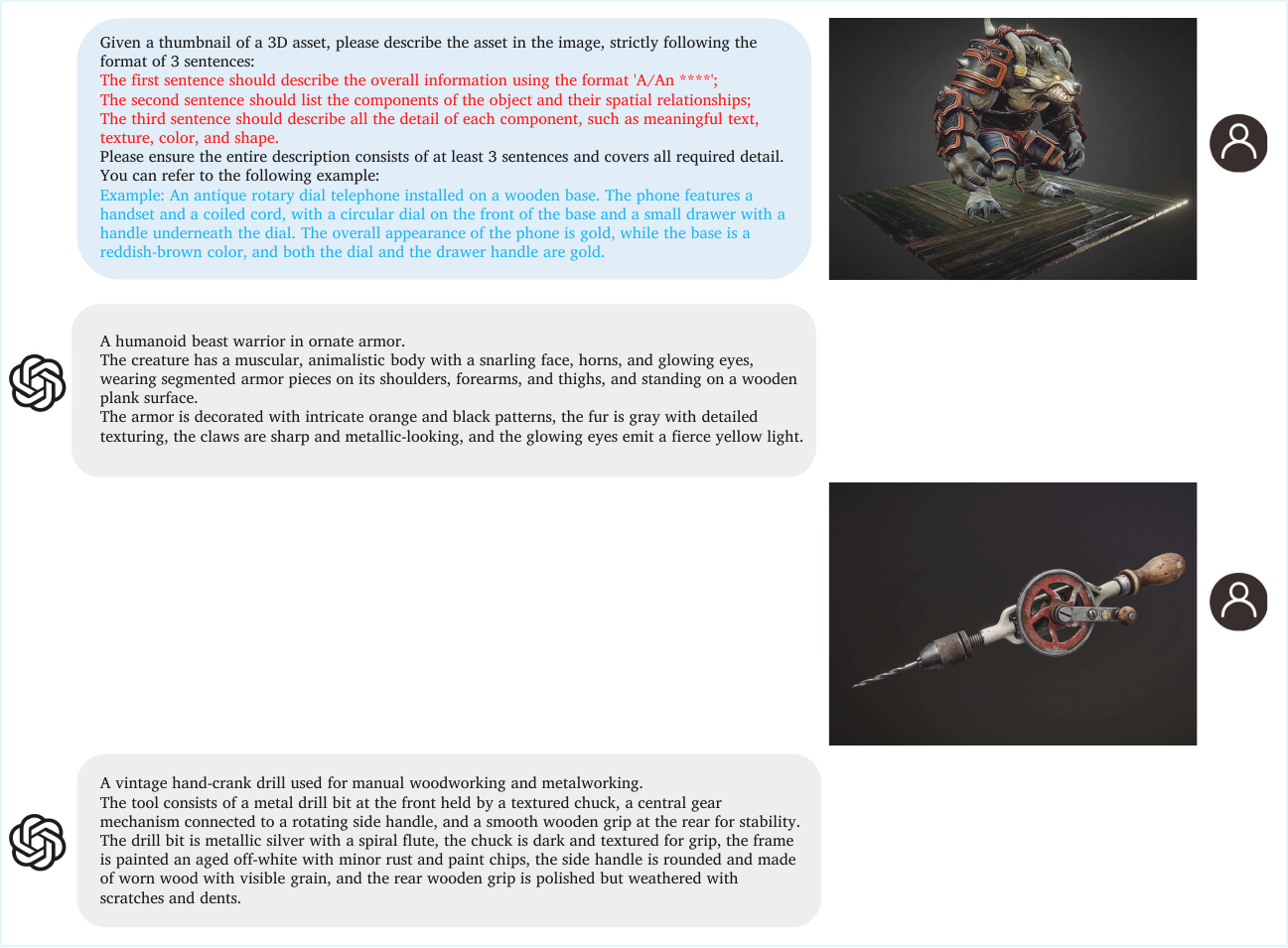}
    \caption{\textbf{Model annotation.} Powered by GPT-5, we provide comprehensive model annotations that capture overall characteristics, structural components, and nuanced features.}
    \label{fig:caption}
\end{figure*}

We provide 856,312 rich model annotations generated by GPT-5 \cite{GPT-5} from the corresponding thumbnails, with the annotation process illustrated in \Cref{fig:caption}.
Following the instruction template inspired by \cite{GT23D-Bench}, GPT-5 is instructed to generate annotations in a fixed three-sentence structure: first, provide an overall description; second, list the components of the object and their spatial relationships; and third, describe each component’s detailed attributes, such as text, texture, color, and shape. 
This standardized format ensures that the resulting descriptions are both comprehensive and consistent.

\section{Conclusion}
We present TexVerse, a large-scale 3D dataset featuring high-resolution textures.
TexVerse includes 858,669 unique 3D models, of which 158,518 incorporate PBR materials.
For rigged and animated models, we preserve the original user-uploaded file in the TexVerse-Skeleton and TexVerse-Animation datasets to maintain skeleton and animation information.
Additionally, we provide 856,312 detailed model annotations generated by GPT-5, encompassing overall descriptions, structural compositions, and intricate feature details.
We believe TexVerse will effectively support future research in fields such as high-resolution texture generation, PBR material synthesis, animation, and a broad spectrum of 3D vision and graphics applications.

\paragraph{Limitations and future work}

We determine model resolution based on metadata from Sketchfab, which may contain occasional annotation errors.
Additionally, while we provide models with high-resolution textures, further filtering and cleaning are needed to ensure consistent geometric quality and texture clarity.
In future work, we will focus on addressing these limitations by developing data validation processes, improving the quality of geometric and texture data through targeted cleaning, and developing more robust annotation methods to enhance dataset reliability.

\clearpage

\clearpage
\bibliographystyle{plain}
\bibliography{main}
\end{document}